\documentclass[]{spie}  

\usepackage{amsmath,amsfonts,amssymb}
\usepackage[colorlinks=true, allcolors=blue]{hyperref}
\usepackage{subfig}
\usepackage{graphicx}
\usepackage{caption}
\graphicspath{ {./} }
 
\renewcommand{\cite}[1]{[\citenum{#1}]}

\title{A Systematic Evaluation of Recent Deep Learning Architectures for Fine-Grained Vehicle Classification}

\author[b,c]{Krassimir Valev}
\author[a]{Arne Schumann}
\author[b,a]{Lars Sommer}
\author[a,b]{J\"urgen Beyerer}
\affil[a]{Fraunhofer IOSB\\
Fraunhoferstrasse 1\\ 76131 Karlsruhe, Germany}
\affil[b]{Vision and Fusion Lab\\
Institute for Anthropomatics and Robotic\\ 
Karlsruhe Institute of Technology KIT\\
Haid-und-Neu-Strasse 7, 76131 Karlsruhe, Germany}
\affil[c]{BMW Car IT GmbH\\
Lise-Meitner-Strasse 14, 89081 Ulm, Germany}

\authorinfo{Further author information: (Send correspondence to arne.schumann@iosb.fraunhofer.de)\\
\{firstname.lastname\}@iosb.fraunhofer.de, Krassimir.Valev@bmw-carit.de
}

\begin{document} 
\maketitle

\begin{abstract}
Fine-grained vehicle classification is the task of classifying make, model, and year of a vehicle.
This is a very challenging task, because vehicles of different types but similar color and viewpoint can often look much more similar than vehicles of same type but differing color and viewpoint.
Vehicle make, model, and year – in combination with vehicle color - are of importance in several applications such as vehicle search, re-identification, tracking, and traffic analysis.
In this work we investigate the suitability of several recent landmark convolutional neural network (CNN) architectures, which have shown top results on large scale image classification tasks, for the task of fine-grained classification of vehicles.
We compare the performance of the networks VGG16, several ResNets, Inception architectures, the recent DenseNets, and MobileNet.
For classification we use the Stanford Cars-196 dataset which features 196 different types of vehicles.
We investigate several aspects of CNN training, such as data augmentation and training from scratch vs. fine-tuning.
Importantly, we introduce no aspects in the architectures or training process which are specific to vehicle classification.
Our final model achieves a state-of-the-art classification accuracy of 94.6\% outperforming all related works, even approaches which are specifically tailored for the task, e.g. by including vehicle part detections.
\end{abstract}


\section{INTRODUCTION}
\label{sec:intro}

Fine-grained vehicle classification refers to the task of recognizing a specific vehicle type, defined by the combination of make, model, and year.
Compared to typical classification problems, such as differentiating cats and dogs or cars and bikes, this fine-grained differentiation between objects of a common basic category is a very difficult task even for humans and generally requires extensive domain knowledge.
Fine-grained vehicle classification finds application in vehicle search and tracking, as well as traffic analysis.

\begin{figure}[ht]
	\centering
    \subfloat[]{{\includegraphics[width=0.30\textwidth]{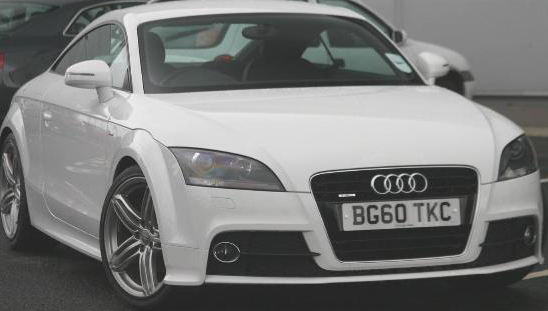} }} 
    \subfloat[]{{\includegraphics[width=0.30\textwidth]{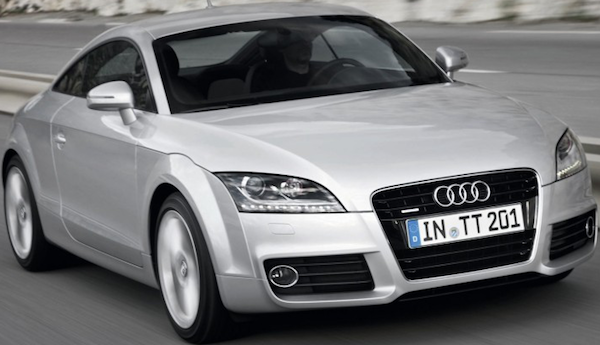} }}
    \subfloat[]{{\includegraphics[width=0.347\textwidth]{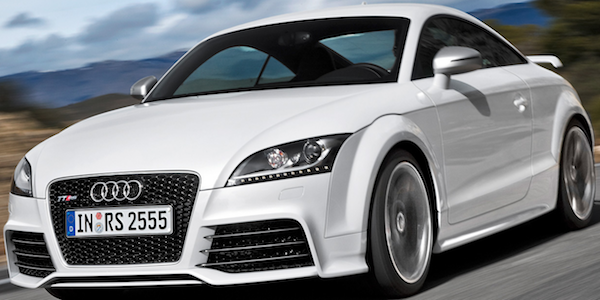} }}\\
    \subfloat[]{{\includegraphics[width=0.32\textwidth]{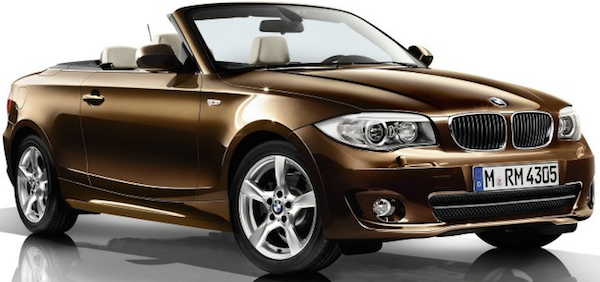} }}
    \subfloat[]{{\includegraphics[width=0.37\textwidth]{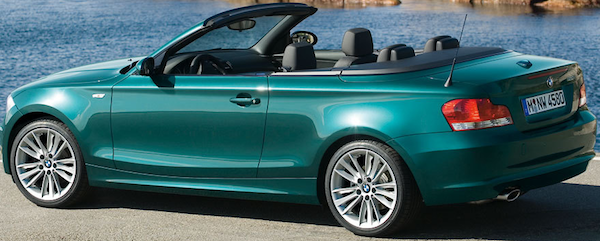} }}
    \subfloat[]{{\includegraphics[width=0.27\textwidth]{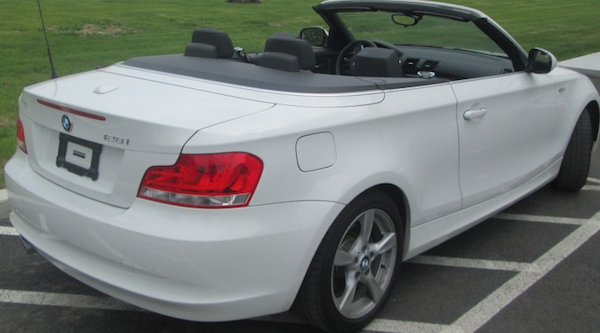} }}
    \caption{Low inter-class variation examples - (a) Audi TTS Coupe 2012, (b) Audi TT Hatchback 2011, (c) Audi TT RS Coupe 2012, (d-f) BMW 1 Series Convertible 2012 \cite{krause20133d}.}
    \label{figure:finegrained_challenges}
\end{figure}

The core challenge of fine-grained vehicle classification is to overcome the high intra-class variance and low inter-class variance that can exist in image space.
Figure \ref{figure:finegrained_challenges} illustrates this problem.
The top row depicts three vehicles with similar viewpoint and color but different combination of make, model, and year.
The visual appearance in the images is very similar, demonstrating a case of low visual variation between classes.
The bottom row, on the other hand, depicts three vehicles of the same make, model, and year but different viewpoints and colors.
The visual variation \emph{within} this class is much greater than the variation \emph{between} classes in the top row.

In order to overcome this challenge, previous work on fine-grained  classification has typically focused on performing part detection in order to overcome the high intra-class variation.
Similarly, 3D representation-based approaches have achieved state-of-the-art results \cite{krause20133d} due to their ability to handle vehicle images with unconstrained poses and multiple viewpoints.
More recently, the success of deep convolutional neural networks (CNNs) has lead to a number of approaches which focus on modifying existing convolutional architectures without explicitly using part annotations \cite{lin2015bilinear, zhao2016diversified, wang2016weakly, hu2017deep}.

In this work we investigate the potential of directly fine-tuning several recent CNN architectures which have shown state-of-the-art accuracy for general image classification tasks.
A variety of experiments are performed with the most promising CNNs in an effort to establish new baselines and inspire future research.
After careful fine-tuning and data augmentation, the baselines presented in this work outperform the previous state-of-the-art by up to 1.5\%.

\section{RELATED WORK}
\label{sec:relwork}

With the successes in large scale image classification in recent years \cite{russakovsky2015imagenet} more and more datasets and approaches address the problem of fine-grained image classification.
This includes approaches and datasets focusing on vehicles \cite{krause20133d}, birds \cite{wah2011birds}, dogs \cite{khosla2011dogs}, and aircraft \cite{maji13aircrafts}.
Our work primarily focuses on the Stanford Cars-196 dataset \cite{krause20133d} which provides class labels for 196 different types of vehicles.
Independent of the basic object category, existing fine-grained classification approaches can be divided into two groups: part-based and CNN-based approaches.

The part-based approach by Wang et al. \cite{wang2010locality} proposes a new coding scheme for traditional spatial pyramid matching (SPM). The Locality-constrained Linear Coding (LLC) scheme uses locality constraints to project each descriptor into its local-coordinate system. Max pooling is used to integrate the projected coordinates and generate the final representation.
The BubbleBank (BB) representation by Deng et al. \cite{deng2013fine} uses crowd-sourced "bubbles" which represent feature templates that are convolved with an image in local search regions, where the regions are determined by the template's image location during training. The final model representation is made up of the pooling responses of a large bank of bubbles over their respective search regions.
Liao et al. \cite{liao2015exploiting} propose a Deformable Part Model (DPM) method for car parts localization and classification. The approach uses supervised DPM to categorize frontal car images and integrates discriminative powers of different parts into the classification.
Krause et al. \cite{krause20133d} lift existing 2D representation approaches (SPM and BB) to 3D with respect to the appearance and location of local features and demonstrate that the resulting 3D object representations outperform both their respective 2D counterparts and state-of-the-art baselines in fine-grained categorization.

Many recent works have shown that the use of CNNs can lead to considerable improvements in general object classification tasks  \cite{krizhevsky2012imagenet,Simonyan14c,He2015}. These CNNs are trained on large datasets such as ImageNet \cite{deng2009imagenet} and it has been shown that the trained convolutional models achieve impressive results for other recognition tasks when used as an image feature extractor \cite{yosinski2014transferable}.
Liu et al. \cite{liumonza} and Yang et al. \cite{yang2015large} investigated using few of the very first deep learning models for the fine-grained car classification task. Their GoogLeNet model was able to surpass the performance of some of the traditional, part-based approaches, reinforcing the belief that using deep convolutional neural networks for fine-grained classification problems is a viable approach, even if it did not not better the state-of-the-art results at the time.
The bilinear model proposed by Lin et al. \cite{lin2015bilinear} consists of two feature extractors whose outputs are multiplied using outer product and pooled to obtain an image descriptor. This architecture models local pairwise feature interactions in a translation-invariant manner making it particularly suitable for fine-grained categorization.
The approach used by Anderson et al. \cite{anderson2016modular} is treating network layers, or entire networks, as modules and combines pretrained modules with untrained modules, to learn the shift in distributions between data sets.
Zhao et al. \cite{zhao2016diversified} have proposed a diversified visual attention network (DVAN), which uses multiple attention canvases to extract convolutional features for attention. An LSTM recurrent unit learns the attentiveness and discrimination of attention canvases.
Hu et al. \cite{hu2017deep} propose a spatially weighted pooling (swp) strategy which aims to considerably improve the robustness and effectiveness of the feature representation of a deep convolutional neural network. The swp uses a predefined number of spatially weighted masks to pool the extracted features of a deep convolutional neural network<<.

\section{METHODOLOGY}
\label{sec:methods}

In this section we first briefly introduce several recent landmark CNN architectures and detail their novel architecture aspects in relation to fine-grained vehicle classification.
We then discuss aspects of training and adapting these models for the vehicle classification task.

\subsection{Convolutional Network Architectures}
\label{sec:architectures}
Neural networks have been used in computer vision for a long time, but with the progress in hardware capabilities and growth of available training data over the last few years deep neural networks have become the most successful methods for many computer vision tasks \cite{krizhevsky2012imagenet}.
In some visual recognition tasks even human-level accuracy can be surpassed \cite{taigman2014deepface,he2015delving}.

The ImageNet Large Scale Visual Recognition Challenge (ILSVRC), started in 2010, has become the established benchmark for large-scale object recognition.
The challenge is based on the publicly available ImageNet dataset \cite{deng2009imagenet}, which provides more than a million training images which are annotated for 1000 different categories.
Over the last five years many deep learning approaches have addressed this challenge and the corresponding models have been published.
Classes in the challenge range from different species of animals over furniture items to classes of vehicles, air-, and watercraft.
Due to this high diversity it is often the case that models which perform well in the ILSVRC challenge can be successfully adapted to other, more specialized computer vision tasks.
We investigate the most prominent and successful architectures published in the ILSVRC challenge and evaluate their performance for the task of fine-grained vehicle classification.


\begin{figure}[ht]
	\centering
    \subfloat[]{{\includegraphics[width=0.25\textwidth]{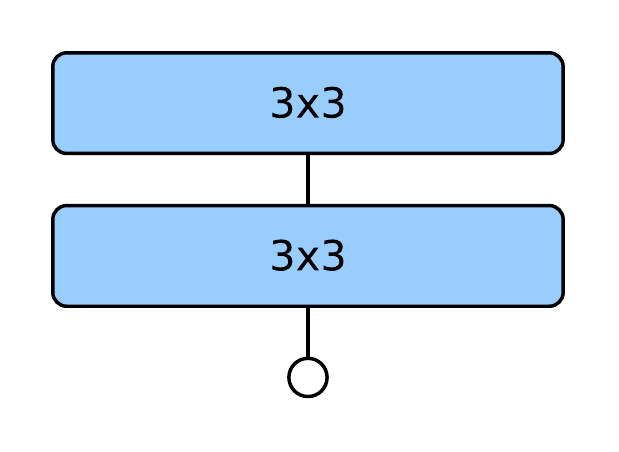} }}
    \hspace{1cm}
    \subfloat[]{{\includegraphics[width=0.25\textwidth]{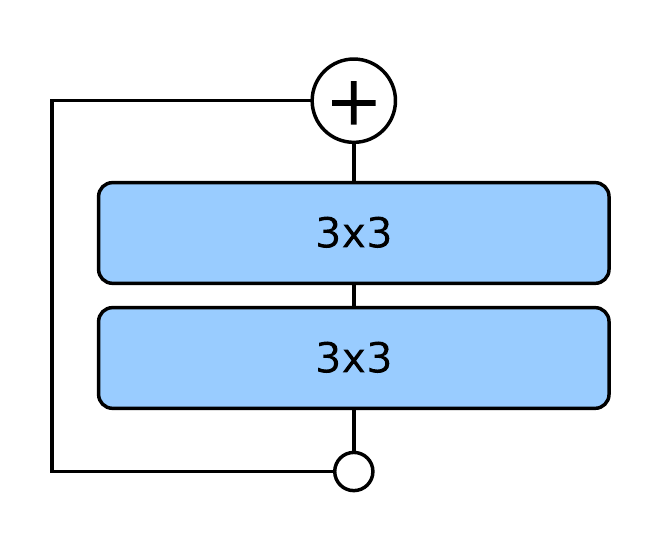} }}
    \hspace{1cm}
    \subfloat[]{{\includegraphics[width=0.25\textwidth]{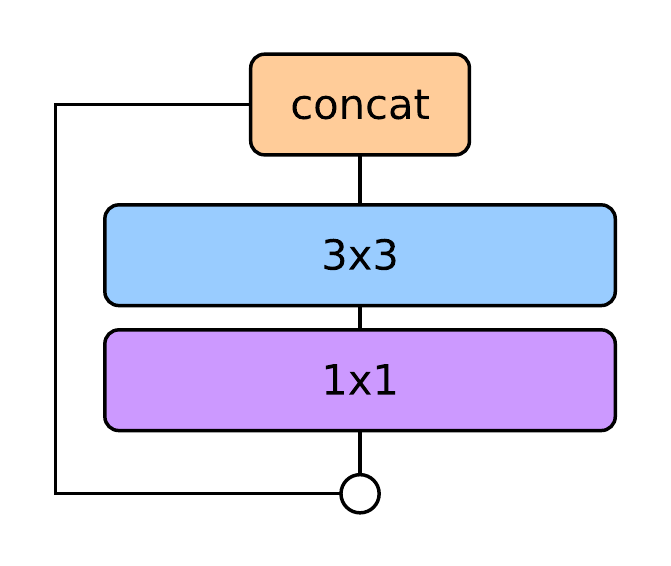} }}
    \caption{Basic building blocks of (a) the VGG architecture \cite{Simonyan14c}, (b) the ResNet architecture \cite{He2015}, and (c) the DenseNet architecture \cite{huang2017densely}.}
    \label{fig:architectures1}
\end{figure}

\subsubsection{VGG (2014)}
After the initial success of the AlexNet architecture \cite{krizhevsky2012imagenet}, the VGG networks \cite{Simonyan14c} were the first to rely strongly on smaller convolutional filters (see Figure \ref{fig:architectures1} (a)).
A sequence of 3$\times$3 convolutions can result in the same receptive field as a single larger convolution while also saving parameters.
For example two 3$\times$3 convolutions have the same receptive field as a single 5$\times$5 convolution and use fewer parameters.
The VGG architectures apply many such 3$\times$3 sequences resulting in a single-path network with a very large number of parameters.

Several different architecture configurations were investigated, the most promising ones being the 16 and 19 layer variants, which achieved a top-5 validation error of respectively 7.5\% and 7.3\% on the ILSVRC challenge.

While the VGG architecture achieved better results than previous approaches, the resulting models have large numbers of parameters which  make a successful training on smaller datasets for fine-grained classification challenging.

\subsubsection{Residual Networks (2015)}
A Residual Network \cite{He2015}, shortened ResNet, is a neural network architecture which solves the problem of vanishing gradients by providing 
elements of the network with an identity shortcut which helps to robustly backpropagate the gradient signal (see Figure \ref{fig:architectures1} (b)).
By stacking these elements, the gradient could theoretically "skip" over all intermediate layers and reach the first layers of the network without being diminished.
These shortcuts also help to simplify the learning task, because the ResNet element only has to learn an offset to its input, due to the addition operation at the end, and no longer a full feature transformation.
The shortcuts thus allow for a successful and robust training of much deeper architectures than previously possible.
The ResNet architecture won the ILSVRC 2015 competition by combining six models of different depths to form an ensemble, achieving an error rate of 3.6\%.

Convolutions of size 1$\times$1 can be applied to reduce the number of channels inside a network element which allows for training of even deeper networks with a reasonable number of parameters.
Such 1$\times$1 convolutions are often called bottleneck layers.

Compared to the VGG model, a more stable gradient should also benefit the fine-grained vehicle classification training of these models.

\subsubsection{Dense Networks (2016)}

Following the idea of identity shortcuts for a more robust training, Dense Networks (DenseNets) \cite{huang2017densely} connect every layer to every other layer inside a block, but the element-wise addition used in ResNets is replaced by a concatenation operation which maintains the individual information of both the input and the skipped layers (see Figure \ref{fig:architectures1} (c)).
This way there is always a direct route for the information backwards through the network.

Similar to other state-of-the-art approaches, DenseNets use 1$\times$1 convolutions as bottlenecks before each 3$\times$3 convolution to reduce the number of input feature maps and to improve computational efficiency.
The concatenation operations result in a strong increase in the number of channels, so a compression approach to reduce the number of channels is applied at transition layers.

The DenseNet results for ILSVRC show that dense architectures perform on par with the state-of-the-art ResNets, while requiring significantly fewer parameters and computation to achieve comparable performance. For example, a DenseNet-201 with 20 million parameters yields similar validation error rate as a 101-layer ResNet with more than 40 million parameters \cite{huang2017densely}.

The smaller number of parameters required by this model is a promising property for training of fine-grained vehicle classification where only comparatively small amounts of training data are available.
Similar to ResNets, the shortcut connections should also help provide a more stable gradient for adaptation to vehicle classification.
In addition to ResNets, the concatenation operation after shortcut connections might better preserve information about image details from earlier layers than an element-wise addition can.
Such details are often crucial for successful fine-grained visual classification.

\begin{figure}[ht]
	\centering
    \subfloat[]{{\includegraphics[width=0.35\textwidth]{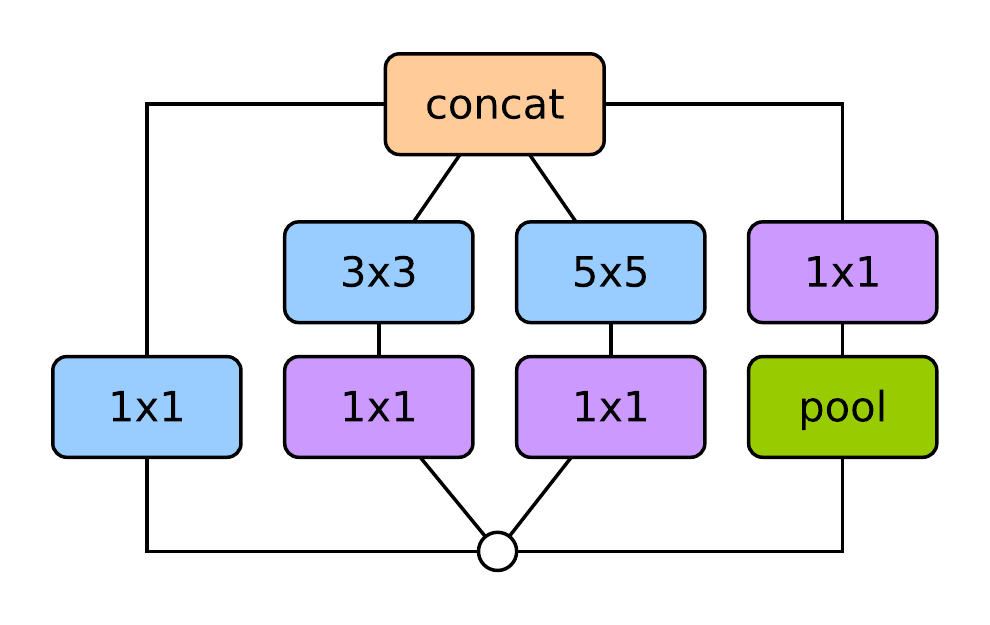} }} \hspace{2cm}
    \subfloat[]{{\includegraphics[width=0.3\textwidth]{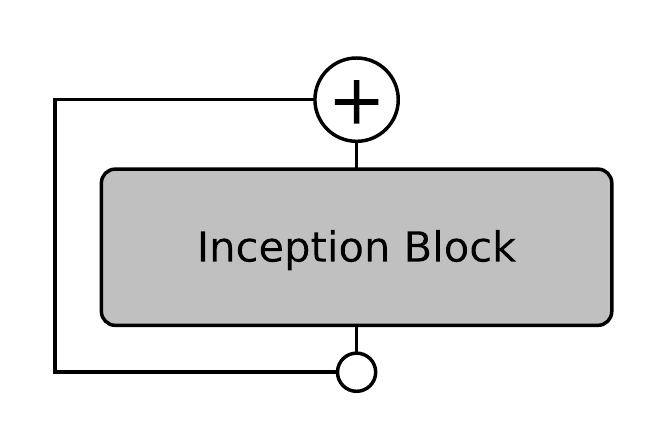} }}
    \caption{(a) An inception block as used in the GoogLeNet architecture \cite{szegedy2015going} and (b) a building block of the Inception-ResNet architecture \cite{szegedy2017inception}.}
    \label{fig:architectures2}
\end{figure}

\subsubsection{Inception ResNet v2 and Inception v4 (2016)}
All inception-based architectures stem from the GoogLeNet model \cite{szegedy2015going}, which won the ILSVRC challenge in 2014 with a top-5 error rate of 6.7\%.
It uses significantly less parameters than other established architectures, while being more accurate.
GoogLeNet was the first architecture that departed from the traditional approach of single-path networks which simply stack convolutional and pooling layers in a sequential structure.
Instead, it uses a deep network element, termed inception module, which was first proposed by Lin et al. \cite{lin2013network} to help enhance model discriminability for local patches within the receptive field. 

The inception module consists of parallel paths of 1$\times$1, 3$\times$3, and 5$\times$5 convolutional filters which are combined through concatenation (as depicted in Figure \ref{fig:architectures2} (a)).
Each branch results in a different receptive field.
This multi-scale view on the input of the module allows the model to recover both local features via smaller convolutions and features with more context through larger convolutions.
Bottleneck layers are used to reduce the number of channels before more complex convolutions.

The latest iteration of the inception network \cite{szegedy2017inception} successfully integrates residual connections and inception modules.
This allows the inception modules to reap the benefits of the residual approach while retaining their computational efficiency.
Furthermore training with residual connections significantly accelerates the training of the resulting Inception-ResNet-v2 architecture. 

In addition to the residual unit-inspired inception architecture, further study has been undertaken on whether the inception architecture itself can be improved by making it deeper and wider, increasing the number of inception modules and simplifying the architecture.
These modifications are combined into the Inception-v4 architecture \cite{szegedy2017inception}.
Similar to previous versions, the newest inception architectures do not use convolutions larger than 3$\times$3, and use factorization to replace large 7$\times$7 filters with a pair of 1$\times$7 and 7$\times$1 convolutional layers.

An ensemble of three residual and one Inception-v4 networks has achieved a 3.1\% top-5 error rate on the ILSVRC classification challenge.

\subsubsection{MobileNet-v1 (2017)}
MobileNet \cite{howard2017mobilenets} is a recent class of deep learning architectures, which is specifically optimized for fast inference on mobile devices.
The main difference between MobileNets and other traditional models is that the standard convolutional operation is decomposed into two more efficient steps.
A depthwise convolutional operation first applies a single filter to each input channel.
A follow-up pointwise convolution then applies a 1$\times$1 convolution to combine the outputs of the depthwise operation.
This factorization drastically reduces the computation time and model size.

Two model hyperparameters can be used to control the trade-off between latency and accuracy.
The width multiplier uniformly reduces the number of channels at each layer.
The resolution multiplier scales down the size of the input image and subsequently all other layers.

MobileNet-v1 reported a 10.5\% top-5 error rate on the ILSVRC challenge.

Given the ILSVRC classification accuracy, it is unlikely that MobileNet will perform as well as other networks when adapted to fine-grained vehicle classification.
However, due to its small model size, the resulting trade-off between accuracy and model size might be helpful when computational power is limited.

\subsection{Transfer Learning and Data Augmentation}
\label{subsec:transfer_learning}

Transfer learning is a machine learning technique that focuses on re-purposing learned classifiers for new tasks.
ImageNet classification models are often a suitable basis for transfer learning, because the classification task they are trained for covers many aspects of the visual world.
This knowledge, encoded in the weights of the model, can serve as a more suitable starting point for many computer vision tasks than, for example, random initialization.
Especially when the amount of training data for a new task is limited, a good initialization becomes crucial to efficiently train large models.

One option to leverage the information of an ImageNet pre-trained model is to use it as a feature extractor. For this, the last fully connected classification layer is removed and the rest of the network is treated as a fixed feature extractor for the new task.
Once the features are extracted on the target dataset, a classifier for the new task can be trained (e.g. a support vector machine or a soft-max layer).

The second and much more popular strategy is to not just replace and retrain the final classifier stage of the convolutional network, but to also fine-tune the weights of the pre-trained network by continuing the backpropagation.
It is possible to fine-tune all the layers of the network or alternatively keep some of the first layers of the network fixed and only fine-tune the higher-level layers of the network.
When fine-tuning a convolutional network it is common to use smaller learning rates for the pre-trained network layers in order to avoid destruction of the initial weights.
Larger learning rates can be used for the randomly initialized classification layer.

\begin{figure}
	\centering
    \subfloat[normal]{{\includegraphics[width=0.19\textwidth]{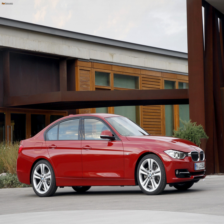} }}
    \subfloat[flip]{{\includegraphics[width=0.19\textwidth]{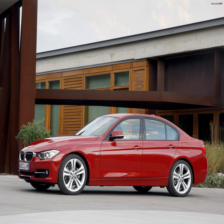} }}
    \subfloat[rotate]{{\includegraphics[width=0.19\textwidth]{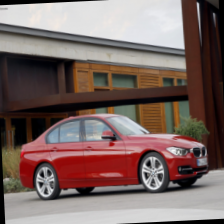} }}
    \subfloat[blur]{{\includegraphics[width=0.19\textwidth]{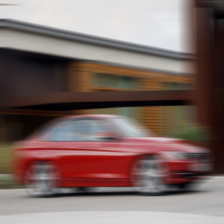} }}
    \subfloat[noise]{{\includegraphics[width=0.19\textwidth]{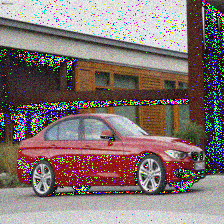} }}
    \caption{Example of several data augmentation strategies. Note that the blur and noise values have been exaggerated for visualization purposes.}
    \label{figure:augmentation_examples}
\end{figure}

Compared to the 1.2 million test images in the ILSVRC training set fine-grained visual recognition datasets are often much smaller.
Since the images for fine-grained recognition are from a much more narrowly defined domain, it is more difficult to gather and annotate similarly large numbers of images.
More common are at most a few hundred classes with some dozens of images per class, resulting in training set sizes of well below 100,000 images.
In order to increase the overall number of training data, but particularly also the number of images within each class, data augmentation strategies can be applied.

The most common strategy is a horizontal flipping of the images.
Besides this, slight rotations can account for unsteady cameras at test time.
Application of motion blur to the images simulates effects resulting from fast motion of vehicles and the addition of image noise can introduce robustness to low image quality into the resulting model.

\section{EVALUATION}
\label{sec:eval}

In this section, we extensively evaluate the performance of the architectures and training configurations described in the previous section on the Stanford Cars-196 dataset \cite{krause20133d}. 
The performance of the newest deep learning models is compared to existing state-of-the-art approaches and new baselines are established.
Our experiments are evaluated based on the reported test dataset top-1 accuracy metric, which compares whether the class predicted with the highest probability matches the ground truth label.

\begin{figure}[h]
	\includegraphics[width=\textwidth]{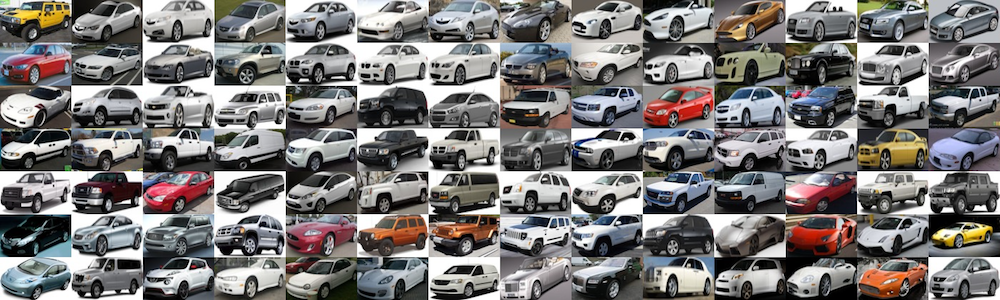}
	\centering
	\caption{Collection of images from the Cars dataset \cite{krause20133d}.}
    \label{fig:dataset}
\end{figure}

The Stanford Cars-196 \cite{krause20133d} dataset contains 16,185 images of 196 car models and is split into a training set of 8,144 images and a testing set of 8,041 images.
Thus, each class has approximately 40 training images and 40 test images.
Since the images contain large background areas, the dataset provides ground-truth annotations of bounding boxes for both the training and test set.
Our experiments are carried out on these bounding boxes.
Some visual impressions of cropped bounding boxes from the dataset are given in Figure \ref{fig:dataset}.

\subsection{Pre-trained vs from scratch}
There are several different approaches when training deep CNN models, depending on the size of the target dataset.
The intuition is that, given the relatively small Stanford Cars-196 training dataset, training a network from scratch will not yield very accurate results.
However, as seen in Table \ref{table:results_finetunevsscratch}, a relatively small 50-layer ResNet cen be successfully trained and achieves quite accurate results (albeit worse than its fine-tuned counterpart).
For larger networks the difference becomes more apparent.
The very deep ResNet-152 has so many parameters that, starting from a random initialization of the weights, the small amount of training data is not sufficient to arrive at a meaningful point in the objective function.

Through fine-tuning, i.e. initialization with ImageNet weights and training at a lower learning rate, both ResNet architectures can be trained successfully and the deeper variant shows a further improvement in accuracy.

\begin{table}[h!]
	\centering
	\begin{tabular}{c|c|c} 
		Architecture & Approach & Accuracy \\
		\hline
		ResNet-50 & from scratch & 84.3\% \\ 
		ResNet-50 & fine-tune & \textbf{92.0\%} \\ 
		ResNet-152 & from scratch & 35.3\% \\ 
		ResNet-152 & fine-tune & \textbf{92.6\%} \\ 
	\end{tabular}
	\caption{Differences between fine-tuning a model and training from scratch}
	\label{table:results_finetunevsscratch}
\end{table}

In lieu of the established practices in \ref{subsec:transfer_learning}, the learning rates for the randomly initialized models were set to a higher value (0.1 compared to 0.01) and a different learning rate step configuration with more iterations was used in order to give the model enough time to converge.

The fine-tuning experiments use the established strategy of replacing the last classification layer with one that matches the number of classes in the new target dataset.
The weights of the new classification layer were randomly initialized using the Xavier algorithm \cite{pmlr-v9-glorot10a}.
Furthermore, none of the pre-trained network layers were frozen, to allow the networks to adapt to the new target dataset.
Experiments with frozen layers were carried out, but showed no benefit.

\subsection{Augmentations}

In Table \ref{table:results_augmentations} we evaluate the impact of common data augmentation methods on fine-grained vehicle recognition.
Using the VGG model as a baseline, simple data augmentation through horizontal flipping can significantly increase accuracy by 2.9\%.
Intuitively, horizontal flipping is one of the most important augmentation methods, because there is no general restriction on which side a vehicle is recorded from and any accidental biases towards this in the training data are reliably eliminated through this augmentation.
The addition of rotation has no meaningful impact on classification accuracy, likely because bounding boxes in the dataset are very well aligned.
Augmentation through adding image noise leads to only slight improvements, because the general image quality of the dataset is very high.
Finally, the addition of motion blur results in the greatest impact over just flipping alone, increasing accuracy by a further 0.2\%.

While rotation, blur, and noise do not result very significant further improvements on this dataset, they are likely to have a positive impact, if the model were to be evaluated in a real world scenario, e.g. on images of moving vehicles recorded by a mobile phone.

\begin{table}[h!]
	\centering
	\begin{tabular}{c|c|c} 
		Model & Augmentations & Accuracy \\ 
		\hline
		VGG16 & - & 83.7\% \\ 
		VGG16 & Flip & 86.6\% \\ 
		VGG16 & Flip, Rotation & 86.6\% \\
		VGG16 & Flip, Noise & 86.7\% \\ 
		VGG16 & Flip, Noise, Rotation & 86.7\% \\
		VGG16 & Flip, Motion blur & \textbf{86.8\%} \\
	\end{tabular}
	\caption{Augmentation strategies effect on model accuracy}
	\label{table:results_augmentations}
\end{table}

\subsection{Architectures \& Comparison to State-of-the-Art}
\begin{figure}
  \begin{minipage}[b]{0.49\textwidth}
\centering
 \begin{tabular}{c|c} 
 	Model & Accuracy \\ 
 	\hline
    CoSeg \cite{krause2015fine} & 92.8\% \\
    VGG-BGL$_m$ \cite{zhou2016fine} & 90.5\% \\
	B-CNN \cite{lin2015bilinear} & 91.3\% \\
    BoT(CNN With Geo) \cite{wang2016mining} & 92.5\% \\
	ResNet50-swp \cite{hu2017deep} & 92.3\% \\
	ResNet101-swp \cite{hu2017deep} & 93.1\% \\
    \hline
    VGG					& 86.8\% \\
    ResNet-50			& 92.0\% \\
	ResNet-152 			& 92.6\% \\
	DenseNet-121 		& 94.0\% \\
	DenseNet-161 		& \textbf{94.6\%} \\
	Inception-v4 		& 90.5\% \\
	Inception-ResNet-v2 & 91.3\% \\
	MobileNet-v1        & 87.6\% \\
\end{tabular}
\captionof{table}{Current state-of-the-art comparison on the Stanford Cars-196 dataset.}
\label{table:stateoftheart_comparison}
  \end{minipage}
  \hfill
  \begin{minipage}[b]{0.49\textwidth}
\centering
  \includegraphics[width=\linewidth]{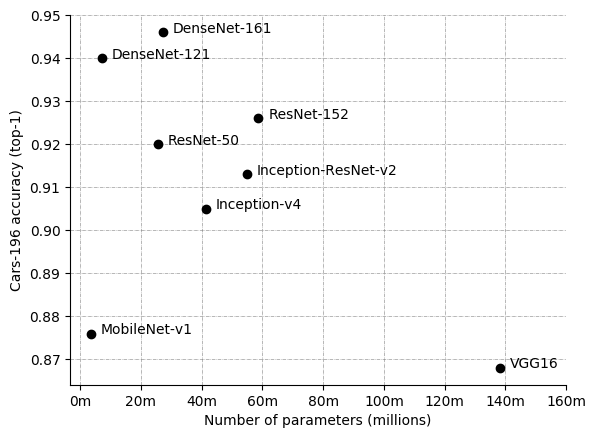}
  \captionof{figure}{Trade-off between number of parameters and model accuracy.}
  \label{fig:tradeoff}
    \end{minipage}
  \end{figure}

In this section we evaluate the accuracies of the different CNN architectures described in Section \ref{sec:architectures} and compare them to specialized state-of-the-art methods on fine-grained vehicle classification. Results are given in Table \ref{table:stateoftheart_comparison}.

It can be observed that the VGG model performs by far the worst.
This is unsurprising, because this model contains no multi-scale representations that preserve details, like inception models, or gradient-stabilizing shortcut connections like ResNets and DenseNets.
The inception models perform better, but are outperformed by the shortcut connection models.
This indicates that shortcuts and a stable gradient are more important for fine-grained vehicle recognition than the multi-scale aspects that inception modules offer.
The purer shortcuts of the DenseNet architectures which do not fuse previous information through addition but maintain it through concatenation with new outputs lead to the highest accuracy.
A reason for this might be that such concatenation-based shortcuts are better able to transport information on vehicle details from early layers through to the classification stage of the network.
The MobileNet model performs similarly weak as the VGG model but requires only a fraction of the parameters.

In Figure \ref{fig:tradeoff} we depict the trade-off between classification accuracy and model complexity.
The results show that the DenseNet models not only outperform other baselines in terms of accuracy but are also among the models with the fewest parameters.
Particularly the DenseNet-121 has almost as few parameters as the MobileNet, which was specifically designed for mobile application, but achieves a 6.4\% higher accuracy than MobileNet.

Interestingly, our fine-tuned DenseNet models which possess no task-specific adaptations outperform the recent state-of-the-art on fine-grained vehicle classification.
Our accuracy of 94.6\% is an increase of 1.5\% over the previous state-of-the-art which corresponds to a 32\% reduction in the remaining classification error.
In combination with specialized adaptations further improvements in accuracy can be expected.

\subsection{Qualitative Results}
\begin{figure}[ht]
\includegraphics[width=\textwidth]{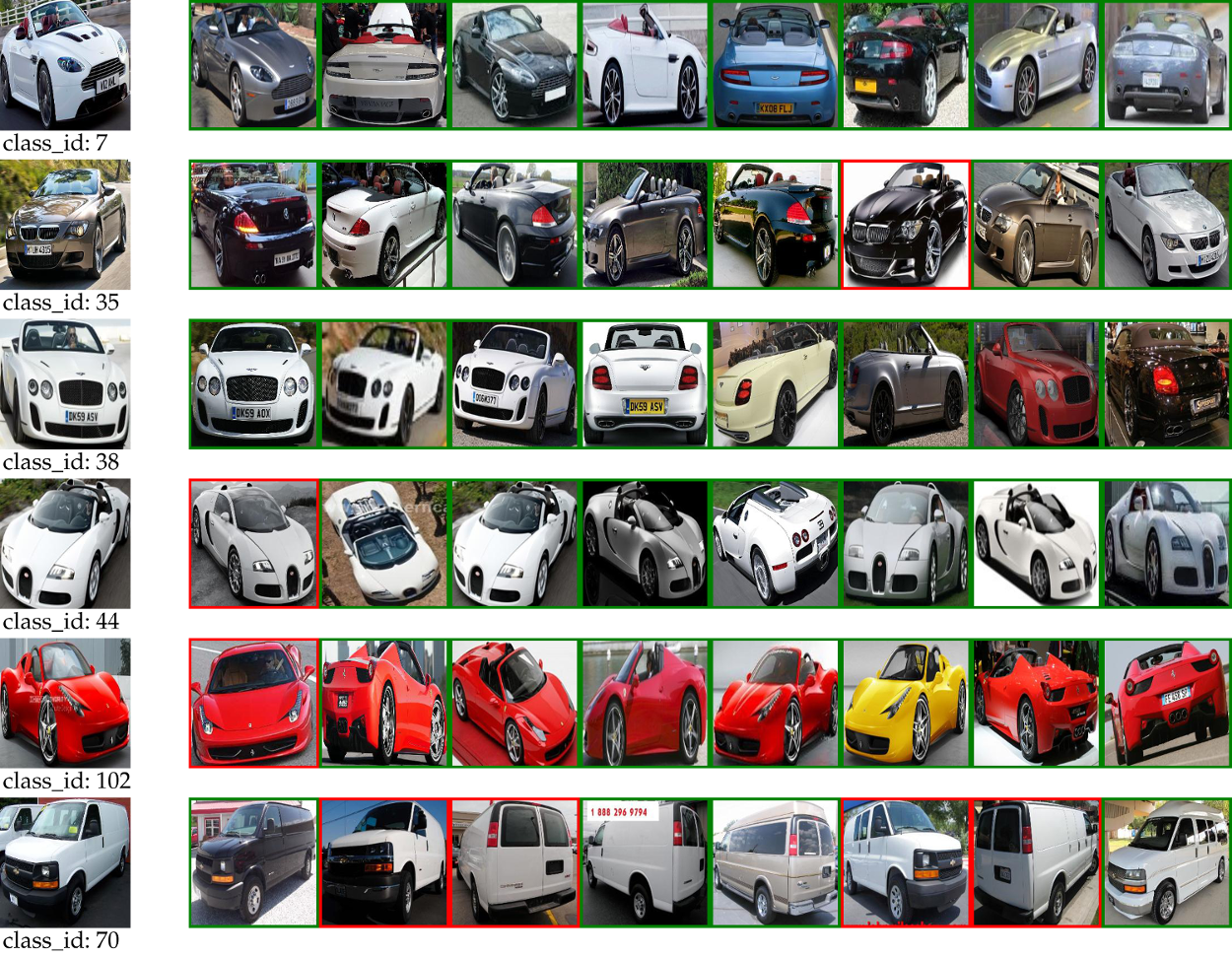}
\centering
\caption{Qualitative results of our DenseNet-161 model. The leftmost image in each row shows an example of the relevant class and the remainder of the row depicts a random sampling of classified examples with correct classification results framed in green and errors framed in red.}
\label{figure:results_showcase}
\end{figure}

Figure \ref{figure:results_showcase} showcases randomly chosen images classified by the DenseNet-161 model.
On the left hand-side a representative image of the given class is shown and on the right-hand side are some of the images, which were categorized to belong to this class.
A green border indicates that the image has been correctly categorized, a red border indicates that the predicted car model does not match the ground truth information.
While most of the car models are classified correctly, there are several class pairs which are extremely similar.
One such case (class\_id: 70) is depicted in the bottom row of Figure \ref{figure:results_showcase}.

\section{CONCLUSION}

With CNN models achieving state-of-the-art results for fine-grained classification tasks, it becomes increasingly important to provide mature and well documented baselines to compare new approaches to.
This work investigates recent landmark CNN architectures on the Stanford Cars-196 dataset.
Due to the small size of the dataset, it was shown that training a CNN from scratch will not yield satisfactory accuracy.
Fine-tuning of existing state-of-the-art architectures trained on ImageNet leads to much better results which could be further improved through data augmentation.
Besides establishing baselines for many important CNN architectures, our DenseNet-161 model even sets a new state-of-the-art of 94.6\% classification accuracy on Stanford Cars-196, outperforming previous approaches by 1.5\%.

\bibliography{report}
\bibliographystyle{spiebib}

\end{document}